\documentclass{article}

\usepackage[accepted]{icml2023}
\usepackage{times}
\usepackage{epsfig}
\usepackage{graphicx}
\usepackage{amsmath}
\usepackage{amssymb}
\usepackage{xfrac}
\usepackage{booktabs,subcaption}
\usepackage{multirow}

\usepackage[pagebackref=true,breaklinks=true,letterpaper=true,colorlinks,bookmarks=false]{hyperref}

\icmltitlerunning{Submission and Formatting Instructions for SPIGM @ICML 2023}

\begin{document}


\twocolumn[
\icmltitle{Empirically Validating Conformal Prediction on Modern Vision Architectures\\
Under Distribution Shift and Long-tailed Data}



\icmlsetsymbol{equal}{*}

\begin{icmlauthorlist}
\icmlauthor{Kevin Kasa}{uofg,vector}
\icmlauthor{Graham W.~Taylor}{uofg,vector}
\end{icmlauthorlist}

\icmlaffiliation{uofg}{School of Engineering, University of Guelph, Guelph, Canada}
\icmlaffiliation{vector}{Vector Institute for Artificial Intelligence, Toronto, Canada}

\icmlcorrespondingauthor{Kevin Kasa}{kkasa@uoguelph.ca}

\icmlkeywords{conformal prediction, uncertainty estimation, out-of-distribution}

\vskip 0.3in
]



\printAffiliationsAndNotice{}

\begin{abstract}
Conformal prediction has emerged as a rigorous means of providing deep learning models with reliable uncertainty estimates and safety guarantees. Yet, its performance is known to degrade under distribution shift and long-tailed class distributions, which are often present in real world applications. Here, we characterize the performance of several post-hoc and training-based conformal prediction methods under these settings, providing the first empirical evaluation on large-scale datasets and models. We show that across numerous conformal methods and neural network families, performance greatly degrades under distribution shifts violating safety guarantees. Similarly, we show that in long-tailed settings the guarantees are frequently violated on many classes. Understanding the limitations of these methods is necessary for deployment in real world and safety-critical applications.
\end{abstract}

\begin{figure}[ht]
\begin{center}
\centerline{\includegraphics[scale=0.75]{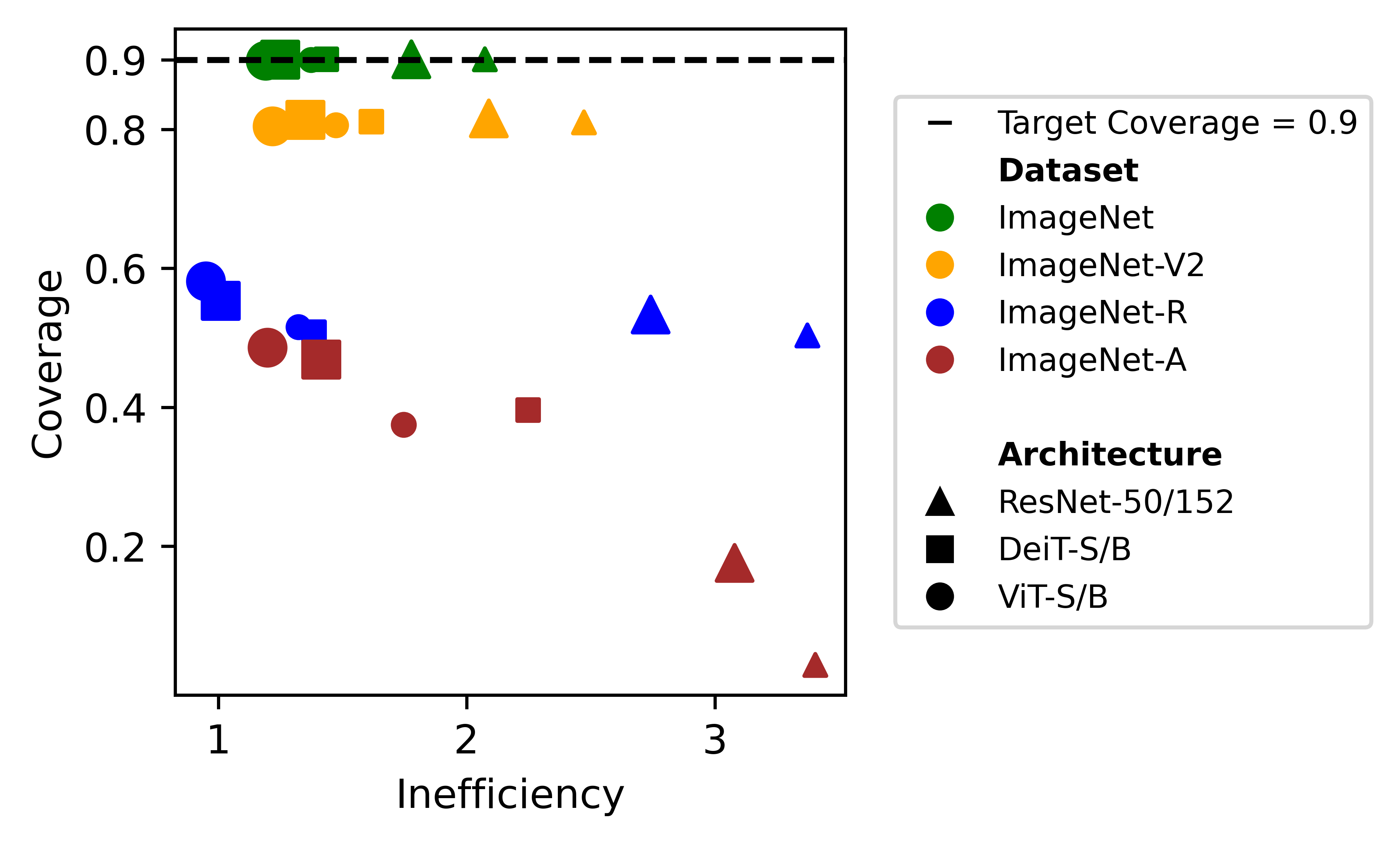}}

\caption{\textbf{Performance of threshold conformal prediction \cite{sadinle_least_2019} degrades across various neural architectures when tested on distribution-shifted ImageNet datasets}.  Target coverage is set to 0.90. All conformal prediction thresholds were first calibrated on a held-out portion of the original validation set. The \textbf{same} threshold was used to construct confidence sets in subsequent test sets. Target coverage is consistently violated for all distribution-shifted sets. Likewise, the average confidence set size, or ``inefficiency'', is observed to increase under distribution shift. Larger markers reflect larger architectures within the family.} 

  \label{fig:ood_IN}
\vskip -0.2in
\end{center}
\end{figure}
\section{Introduction}
Deep learning models have shown the ability to complete a diverse range of tasks with exceedingly high performance \cite{silver2017mastering, LLM, ViT}. However, high performance metrics (e.g., accuracy) alone are insufficient for deployment in safety-critical applications, where uncertainty measures and safety guarantees that experts can trust are required \cite{ovadia2019trust}. \textit{Conformal prediction} (CP) \cite{conformal_prediction_vovk} is a promising method for addressing these limitations. Conformal prediction turns heuristic notions of uncertainty into reliable ones through a post-training adjustment, which can then be used to predict \textit{confidence sets} that are guaranteed to contain the true class with some user specified error rate. \par 

Various conformal prediction methods \cite{sadinle_least_2019, stutz_learning_2022, romano_classification_2020, angelopoulos_uncertainty_2022, teng2023predictive} perform well on a number of complex tasks such as image classification and object detection \cite{angelopoulos_learn_2022}. However, these results thus far are largely limited to in-distribution and class-balanced data regimes. This is problematic since data encountered in real-world settings is often imbalanced \cite{Krawczyk2016LearningFI} or subject to distribution shift \cite{Castro_2020}, and robustness to these settings is necessary for the safe deployment of ML \cite{amodei2016concrete}.  

Despite the importance of understanding performance in these real-world settings, there has thus far been no comprehensive investigation of the performance of popular conformal prediction methods under distribution shift and long-tailed data. Since conformal prediction assumes identically distributed data and guarantees provided are based on micro- rather than macro-averages, it is unsurprising that performance would degrade under shifted and long-tailed distributions. This phenomenon has been observed in small-scale datasets \cite{tibshirani2020conformal}. Nonetheless, the recent adoption of conformal prediction into deep learning and safety-critical domains \cite{angelopoulos_learn_2022, muthali2023multiagent, vazquez_facelli_2022, 10.1007/978-3-031-16452-1_52} warrants specific investigation of these methods using modern neural network architectures and large-scale datasets that are more characteristic of data found ``in the wild''.

In this study, we evaluate four different conformal prediction methods on numerous distribution-shifted and long-tailed datasets and thoroughly characterize their performance under these conditions. We investigate across three deep learning model families, while also controlling for model size. Our primary findings are:
\begin{itemize}
    \item Safety guarantees in terms of coverage (Eq.~\ref{eq8}) are violated even under small distribution shifts. 
    \item Class-conditional coverage is frequently violated in long-tailed distributions.
    \item The size of the confidence sets, with smaller being more desirable, increases under both these settings. 
    \item The above results hold across all CP methods and model architectures. 
\end{itemize}

\section{Methods}
In this study, four conformal prediction methods were evaluated across five distribution-shifted datasets and one long-tailed dataset, for image classification tasks. Three neural architecture families were used as the base classifier, to determine their affect on CP performance, which was evaluated using several metrics. 

\subsection{Conformal Prediction Methods}
The common classification paradigm involves training a model $\pi_\theta(x)$ to predict a \emph{single label} $Y \in [K] := \{1, ..., K\}$. In contrast, conformal prediction is a statistical method that can be used to predict confidence \textit{sets} for machine learning models \cite{angelopoulos_uncertainty_2022}. Formally, it aims to construct a confidence set $\mathcal{C} \subseteq [K]$ such that the true class is included with some user specified error rate $\alpha$:
\begin{equation} \label{eq1}
    \mathbb{P}(Y_\text{test} \in \mathcal{C}(X_\text{test})) \ge 1 - \alpha.
\end{equation}
This is done through a two step post-processing procedure. In the calibration step, a score function $s(x,y)$ is used on held-out data to transform a provisional uncertainty measure (e.g., softmax values) into \textit{conformity scores}. The $1-\alpha$ quantile of the conformity scores is then used to determine a threshold $\tau$. In the prediction step, sets $\mathcal{C}(X)$ are constructed on new unseen data by including all the labels whose conformity scores fall within the threshold, guaranteeing $1-\alpha$ coverage. Importantly, this guarantee is known as \textit{marginal} coverage, since it holds in expectation \emph{unconditionally} across all data points rather than per-class. The returned confidence sets can also be used as an uncertainty estimate, with larger confidence sets $\left| \mathcal{C}(X) \right|$ suggesting greater uncertainty in the predictions. 

The \textbf{threshold conformal prediction (THR)} method \cite{sadinle_least_2019} generally produces the smallest average set sizes. Here, the confidence sets are constructed as:
\begin{equation}\label{eq2}
    \mathcal{C}(x; \tau) := \{k \in [K]: s(x,k) > \tau\}
\end{equation}
Here, the score function is defined as $s(x, y) = \pi_\theta(x)_{y}$, and the threshold $\tau$ is computed as the $\alpha \left(1+\sfrac{1}{N_{\text{cal}}} \right)$ quantile of the calibrated conformity scores. During calibration, the softmax value corresponding to the true class $y$ of the input $x$ is used in the conformity scores. At test time, this method includes in the set those classes whose softmax score is greater than the calibrated threshold. Although THR produces small set sizes, it may lead to ``uneven'' coverage, with difficult classes achieving worse coverage. 

\textbf{Adaptive prediction sets (APS)} \cite{romano_classification_2020} were developed to improve conditional coverage, with the trade-off of larger set sizes.
In the APS method, the conformity scores are calculated by accumulating softmax values:

\begin{equation} \label{eq4}
    s(x,y) = \sum_{j=1}^y \hat{\pi}_{\theta} (x)_j,
\end{equation}

Where $\hat{\pi}_{\theta} (x)$ is the sorted softmax values for input $x$ from greatest to smallest. Subsequently, sets are constructed by including values \textit{less} than the threshold $\tau$:
\begin{equation} \label{eq3}
    \mathcal{C}(x;\tau) := \{k \in [K]: s(x,k) < \tau \},
\end{equation}

Similarly to THR, the conformity scores with respect to the true class $y_i$ are used for calibration, and the $(1-\alpha)(1 + \sfrac{1}{N_{\text{cal}}})$ quantile is used to find the value $\tau$ that ensures marginal coverage on test examples.  

\textbf{Regularized adaptive prediction sets (RAPS)} \cite{angelopoulos_uncertainty_2022} build on APS by modifying the conformity scores to include a penalty $\lambda$ to classes beyond some specified threshold $k_{reg}$. Specifically, the score function is defined as:

\begin{equation}\label{eq5}
    s(x,y) := \sum_{j=1}^k \pi_{\theta} (x)_y + \lambda \cdot (o_x(y) - k_{\text{reg}})^+,
\end{equation}

\noindent where $o_x(y)$ is the ranking of label $y$ among the sorted probabilities, and $\left(\cdot\right)^+$ indicates the positive part of the expression. The confidence sets are then defined the same as in Equation \ref{eq3}. The regularization helps to exclude probabilities that are deep in the tail that would otherwise have been included, since labels now require a greater score to be included in the set. This helps to produce smaller prediction sets than APS (albeit not as small as THR), and has been shown to work well on large datasets like ImageNet \cite{angelopoulos_uncertainty_2022}. In our experiments, convolution-based networks use values of $\lambda=0.01$ and $k_{reg}=5$, and transformer-based networks use $\lambda=0.1$ and $k_{reg}=2$. 


The CP methods described thus far are implemented \textit{after} a model is trained, which does not directly optimize the underlying model to produce high performing confidence sets. \textbf{Conformal training (ConfTr)} \cite{stutz_learning_2022} was proposed to address this, by simulating the conformal prediction process during training. This is done by splitting each training batch $B$ into a calibration $B_{cal}$ and prediction 
$B_{pred}$ subset. Just like in regular CP, $B_{cal}$ is used to calibrate the threshold $\tau$, and confidence sets are formed on $B_{pred}$. To perform the thresholding step, differentiable sorting \cite{blondel2020fast} is used to find the quantiles of the conformity scores in a way that can be back-propagated during training. The size of the confidence sets is then used as the loss function to be minimized during training:

\begin{equation} \label{eq6}
    \mathcal{L}_{\text{size}} = \max \left(0, \sum_{k=1}^K E_{\theta,k}(x;\tau) -\kappa \right).
\end{equation}

\noindent In Equation \ref{eq6}, $E_{\theta,k}(x;\tau)$  is a ``smooth'' assignment of class \textit{k} to the confidence set, calculated as $E_{\theta,k}(x;\tau) := \sigma\left(\frac{s(x,y) - \tau}{T}\right)$, where $\sigma(\cdot)$ is the Sigmoid function and $T \in [0,1]$ is a temperature parameter controlling the smoothness. This penalizes the set sizes, and the hyper-parameter $\kappa \in \{0,1\}$  determines whether or not sets of size one are penalized (i.e., $\kappa=1$ means that singleton sets will incur no loss). An additional classification loss can be included to ensure the true label is included in the confidence sets:
\begin{equation}\label{eq7}
    \mathcal{L}_{\text{class}} = \sum_{k=1}^K\left[(1-C_{\theta, k}(x;\tau)) \cdot \mathbf{1}[y=k] \right].
\end{equation}
A weighted combination $\mathcal{L} = \mathcal{L}_{\text{class}} + \lambda \mathcal{L}_{\text{size}}$ can then be used to train the model.

For this method, a ResNet-50 pre-trained on ImageNet \cite{rw2019timm} was used as the base model. The training methodology and hyper-parameters closely follow that used by the original authors on the CIFAR-100 dataset \cite{stutz_learning_2022}. This included re-initializing the final fully connected layer, and training one baseline model using cross-entropy loss and one with the combined $\mathcal{L}_{\text{size}}$ and $\mathcal{L}_{\text{class}}$ losses, defined in Equation \ref{eq6} and Equation \ref{eq7}.  

Any CP method can be used to predict the confidence sets during training, however in practise THR has been shown to produce better results, so that is used in this study for the ConfTr experiments. Because ConfTr relies on smooth sorting / assignment operations, post-training conformal prediction is still performed to ensure the formal guarantees are maintained.

 
\subsection{Evaluation Metrics}
The primary metrics used for evaluation are coverage and inefficiency. \textbf{Coverage} measures the fraction of true labels that are actually included in the confidence set:
\begin{equation}\label{eq8}
    \textit{Cover} := \frac{1}{N_{\text{test}}} \sum_{i = 1}^{N_{\text{test}}} \mathbf{1} [y_i \in \mathcal{C}(x_i)].
\end{equation}
The conformal prediction process guarantees that $\mathbb{P}(Y_{\text{test}} \in \mathcal{C}(X_{\text{test}})) \ge 1 - \alpha$, thus the $\textit{Cover}$ metric should be $\geq 1-\alpha$ on average. However, conformal prediction does not guarantee \textbf{class conditional coverage}: $\mathbb{P}(Y_{\text{test}} \in \mathcal{C}(X_{\text{test}}) | Y_{\text{test}} = y) \ge 1 - \alpha$. We can capture conditional performance using a ``macro'' coverage metric. First we can consider $\textit{Cover}(k)$ to be the the coverage computed only on test points from class $k \in [K]$. The macro coverage is then:
\begin{equation} \label{macro_cov}
    \textit{Macro Cover} := \frac{1}{K} \sum_{k=1}^K \textit{Cover}(k).
\end{equation}
The non-conditional guarantees of conformal prediction mean that although across an entire dataset the desired coverage may be maintained, there may be classes which violate the desired coverage level. This is especially pertinent for long-tailed datasets. Thus, the number of classes that violates the coverage level is found:
\begin{equation}
    \textit{Cover Violation} :=  \sum_{k=1}^K \mathbf{1}\left[ \textit{Cover}(k) < 1-\alpha \right].
\end{equation}
\textbf{Inefficiency} is a measure of the size of the confidence sets. The prediction sets must both provide adequate coverage (contain the right class), and be \textit{informative}; very large prediction sets are of little use. Inefficiency is measured as:
\begin{equation}
    \textit{Ineff} := \frac{1}{N_{\text{test}}} \sum_{i=1}^{N_{\text{test}}} \left| \mathcal{C}(x_i) \right|.
\end{equation}
The macro inefficiency is also calculated, to determine if some classes tend to return particularly large sets. Similarly to Equation \ref{macro_cov}, we define $\textit{Ineff}(x)$ as the inefficiency on class $k$, and the macro inefficiency as: 
\begin{equation}
    \textit{Macro Ineff} := \frac{1}{K} \sum_{k=1}^K \textit{Ineff}(k).
\end{equation}
The macro coverage and inefficiency metrics will be used to characterize performance on the long-tailed datasets.  

\subsection{Datasets}

\paragraph{Distribution Shift.}

We use the ImageNet \cite{deng2009imagenet} dataset to train our neural networks and calibrate the CP classifiers. Following previous works on conformal prediction \cite{angelopoulos_uncertainty_2022}, we reserve 50\% of the ImageNet validation set to find the threshold $\tau$. This \textbf{same threshold} is used to form prediction sets on the remaining ImageNet validation set, as well as the following distribution-shifted datasets:

\begin{enumerate}
    \item \textbf{ImageNetV2} \cite{Recht2019DoIC} is a new ImageNet test set collected by closely following the same format and collection process as ImageNet, with the goal of mimicking the original data distribution.\footnote{It is difficult to conclude whether this dataset represents a true distribution shift in the absence of convincing generalization error bounds for ImageNet-scale DNNs, however, we adopt~\citeauthor{Recht2019DoIC}'s hypothesis that it indeed represents a small shift.}
    \item \textbf{ImageNet-C} \cite{Hendrycks2018BenchmarkingNN} applies common visual corruptions to the ImageNet validation set. In this study, the Gaussian noise, motion blur, brightness, and contrast corruptions are investigated, representative of the four main categories --- noise, blur, weather, and digital, respectively. 
    \item \textbf{ImageNet-A} \cite{hendrycks2021nae} contains naturally adversarial images that a ResNet-50 incorrectly classifies, but can be correctly classified by humans. 
    \item \textbf{ImageNet-R} \cite{hendrycks2021many} consists of rendered versions of ImageNet classes, such as drawings, cartoons, etc.
\end{enumerate}
The details of these datasets are summarized in Table \ref{tab:ood_data}. Metrics are reported as the average across ten trials, to account for variation in the calibration split. 

\begin{table}[h]
    \caption{Alternate ImageNet-based validation datasets used to evaluate performance under distribution shift. For ImageNet-C, the Gaussian noise, motion blur, brightness, and contrast corruptions are used. The conformal calibration process is \textbf{only} conducted on the original ImageNet validation set.}
    \centering
    \resizebox{\columnwidth}{!}{
    \begin{tabular}{|l|c|c|}
        \hline
         Dataset & Number of Images & Number of Classes \\
         \hline
         ImageNet-V2~\cite{Recht2019DoIC} & 10,000 & 1,000 \\
         ImageNet-C~\cite{Hendrycks2018BenchmarkingNN} & 50,000 & 1,000 \\
         ImageNet-A~\cite{hendrycks2021nae} & 7,500 & 200 \\
         ImageNet-R~\cite{hendrycks2021many} & 30,000 & 200 \\
        
         \hline
    \end{tabular}
    }
    \label{tab:ood_data}
\end{table}

\paragraph{Long-tailed labels.} Conformal prediction performance on long-tailed data distributions was evaluated on the PlantNet-300k dataset \cite{plantnet-300k}. This is a highly imbalanced dataset, with 80\% of classes accounting for only 11\% of the total number of images. In addition to the 243,916 training examples, PlantNet-300k has defined validation and test sets, each with 31,118 examples and at least one image of each class in each set. The validation set is used to calibrate the conformal prediction methods and find the threshold, and the test set is used to form confidence sets and evaluate performance. Here, all three data splits (train, validation, and test) are long-tailed, meaning that \textbf{the conformal calibration process is conducted on highly imbalanced data}. 

\subsection{Deep Learning Models}
To account for differences in model architecture and training algorithms, three distinct model families were evaluated:

\begin{enumerate}
    \item \textbf{ResNets} \cite{He2015DeepRL} are prototypical convolutional neural networks.
    \item \textbf{Vision Transformers (ViT)} \cite{Dosovitskiy2020AnII} are transformer-based architectures that are pre-trained on ImageNet-21k \cite{ridnik2021imagenet21k}, before being fine-tuned on ImageNet-1k. 
    \item \textbf{Data efficient image Transformers (DeiT)} \cite{Touvron2022DeiTIR} are also transformer networks, however they are trained only on ImageNet-1k following a carefully designed training procedure. 
\end{enumerate}

\section{Experiments and Results}
\subsection{Distribution Shift}

\begin{table}[t]
    \caption{Comparison between using baseline cross-entropy training and ConfTr, which directly optimizes set sizes during training. Although ConfTr leads to smaller sizes on the in-distribution test data, there is negligible difference in coverage between the two methods on ImageNet-V2. Coverage target is 0.99.}
\centering
\resizebox{\columnwidth}{!}{

    \begin{tabular}{@{}lcccccc@{}}
    \toprule
      \multirow[c]{2}{*}{Datasets} & \multicolumn{2}{c}{Accuracy}  &   \multicolumn{2}{c}{Coverage}    &  \multicolumn{2}{c}{Inefficiency}  \\
    & Baseline & ConfTr &  Baseline & ConfTr &  Baseline & ConfTr  \\
    \hline
    \addlinespace
    ImageNet & 76.91 & 72.40 & 0.99 & 0.99 & 32.21 & 29.89 \\
    ImageNet-V2 & 64.68 & 60.45 & 0.97 & 0.97 & 50.79 & 46.99 \\
    \bottomrule
    \end{tabular}   
    }
    \label{tab:conftr}
\end{table}

\begin{table*}[t]

\caption{Conformal prediction results on PlantNet-300k. While marginal coverage of 0.90 is maintained, class-conditional coverage is frequently violated. The conformal threshold is calibrated on a (long-tailed) held-out validation set.}

\centering
\scalebox{0.82}{
    \begin{tabular}{@{}lcccccccc@{}}
    \toprule
      Model & CP Method & Accuracy & Macro Acc. & Coverage & Macro Coverage & Inefficiency & Macro Inefficiency & \# (\%)~Violated Classes \\
    \hline
    \addlinespace
     \multirow[c]{3}{*}{ResNet-152} & THR &\multirow[c]{3}{*}{80.84} & \multirow[c]{3}{*}{36.82} & 0.899 & 0.505 & 1.46 & 1.99 & 774~(72\%)\\
     & APS & & & 0.900 & 0.648 & 3.67 & 13.75 & 617~(57\%) \\
     & RAPS & & & 0.900 & 0.610 & 2.15 & 5.17 & 665~(62\%) \\
    \addlinespace
     \multirow[c]{3}{*}{DeiT-B} & THR &\multirow[c]{3}{*}{82.68} & \multirow[c]{3}{*}{43.57} & 0.898 & 0.541 & 1.30 & 1.50 & 714~(66\%) \\
     & APS & & & 0.900 & 0.713 & 4.70 & 18.30 & 513~(47\%) \\
     & RAPS & & & 0.901 & 0.603 & 1.68 & 2.64 & 654~(60\%) \\
    \addlinespace
     \multirow[c]{3}{*}{ViT-B} & THR &\multirow[c]{3}{*}{82.15} & \multirow[c]{3}{*}{35.86} & 0.899 & 0.461 & 1.56 & 2.42 & 800~(74\%) \\
     & APS & & & 0.899 & 0.744 & 12.37 & 98.45 & 466~(43\%) \\
     & RAPS & & & 0.901 & 0.551 & 1.67 & 3.27 & 697~(64\%)\\
     \bottomrule
    \end{tabular}   
}
\label{tab:pnt1}



\label{tab:pn300k}
\end{table*}

Our results on alternate ImageNet test sets are summarized in Figure \ref{fig:ood_IN}. We can see that the desired coverage is consistently violated across all models. Distribution shift also leads to increased inefficiency --- a proxy for the increased uncertainty of the underlying model. The coverage target is violated even on small distribution shifts, such as ImageNet-V2, which was purposefully and carefully constructed to match the original ImageNet distribution as closely as possible. The inability of these methods to maintain coverage even on minor distribution shifts highlights the risks of deployment in real world situations, without additional safety features.  \par
Smaller models exhibit worse inefficiency, and often lower coverage rates. The larger ViT / DeiT models perform best overall with the smallest degradation under distribution shift. These results highlight the value of combining conformal prediction with modern, high-performing deep learning models. It affirms that efforts to improve the performance of the base model may improve the performance of conformal prediction methods under distribution shift. Refer to Appendix \ref{detailed_ood} for detailed results on these datasets, as well as ImageNet-C results at each corruption level. Further, Appendix \ref{apndx:acc_vs_cov} shows the relationship between model accuracy and CP coverage, and Appendix \ref{apndx:INw} includes results on the recent ImageNet-W \cite{li_2023_whac_a_mole} dataset.

Table \ref{tab:conftr} shows the results of the conformal training method. As expected, the ConfTr method leads to smaller sets on the in-distribution data, however, this does not translate to improved coverage on distribution-shifted data. 

\subsection{Long-tailed Label Distributions}
Table \ref{tab:pn300k} shows the results on the long-tailed PlantNet-300k dataset. Although the target coverage of 0.90 is maintained marginally across the entire dataset, it is frequently violated on a class-conditional basis. Indeed, there are often hundreds of classes with violated coverage levels, leading to a violation of coverage on up to 70\% of the classes in the worst case. This is consistent across all models and methods, and highlights the difficulty of applying conformal prediction methods to long-tailed data distributions.\par
The ineffectiveness of approximating class-conditional coverage on PlantNet-300k is further demonstrated in the Appendix (see Table \ref{tab:pn300k-cb}). The Appendix also includes the results of experiments on the iNaturalist-2018 \cite{inaturalist18} and -2019 \cite{inaturalist19} datasets (see Table \ref{tab:inat}).

\section{Conclusion}
In this paper, we studied the performance of conformal prediction methods under distribution shift and long-tailed data, on large-scale datasets and modern neural architectures. We show that performance degrades in these regimes, and coverage guarantees are frequently violated. We also observed increased inefficiency, the average size of the conformal sets. While violation of coverage guarantees is undesirable, inefficiency indicates model uncertainty. A good model should exhibit heightened uncertainty with OOD examples.

There have been several recent methods developed in dealing with distribution shift \cite{amoukou_adaptive_2023, gendler_adversarially_2022, gibbs_conformal_2022, barber_conformal_2023, dunn_distribution-free_2022, bhatnagar_improved_2023, cauchois_knowing_nodate, gibbs2023conformal} and class-conditional coverage \cite{deng_group_2023, fisch_few-shot_2021, jung_batch_2022}. However, these have thus far been developed mostly on small-scale datasets, and it remains to be seen how they translate to the large-scale datasets studied here. This is something future works may tackle, and we hope that our results will serve as baselines upon which new conformal prediction methods and novel algorithms and architectures for deep learning can improve.  

Ultimately, this work highlights the challenges that conformal prediction methods may face in real world applications, where class imbalance is common and data distributions are ever-shifting. Developing and empirically evaluating conformal prediction methods that are more robust to these admittedly difficult settings is a key requirement to their adoption in safety-critical environments. 

\bibliographystyle{icml2023}

\newpage
\appendix
\onecolumn

\section{Detailed Results on ImageNet Distribution Shift} \label{detailed_ood}

The detailed results on alternate ImageNet test sets are reported in Table \ref{tab:IN-OOD}. As mentioned, the target coverage level of 0.90 is violated in nearly all circumstances. We see that THR indeed provides the smallest set sizes, while the adaptability of RAPS generally results in better, but imperfect, coverage. Further, the basic APS method often leads to impractically large set sizes. In addition to degrading coverage, the inefficiency also increases on these datasets; a proxy for the increased uncertainty of the underlying model. 


        


\begin{table*}[htbp]

\caption{Performance of various conformal prediction and neural architectures on distribution-shifted ImageNet datasets with a target coverage of 0.90. All conformal prediction thresholds were first calibrated on a held-out portion of the original validation set. The \textbf{same} threshold was used for constructing confidence sets in the subsequent test sets. Target coverage is consistently violated for all distribution-shifted sets. Likewise, the average confidence set size, a.k.a.~``inefficiency'', is observed to increase under distribution shift.} \label{tab:overall}

\begin{subtable}{0.5\textwidth}
\caption{Performance on original ImageNet validation set}\label{tab:INval}
\centering
\scalebox{0.75}{
\begin{tabular}{@{}lccccccc@{}}
\toprule
  \multirow[c]{2}{*}{Model} & \multirow[c]{2}{*}{Accuracy}  &   \multicolumn{3}{c}{Coverage}    &  \multicolumn{3}{c}{Inefficiency}  \\
& & THR & APS & RAPS & THR & APS & RAPS  \\
\hline
\addlinespace
ResNet-50 & 76.14& 0.899 & 0.899 & 0.899 &  2.05 & 9.06 & 3.78 \\
ResNet-152 & 78.04 & 0.900 & 0.900 & 0.900 & 1.79 & 6.37 & 2.98 \\
\addlinespace
DeiT-S & 81.29 & 0.900 & 0.900 & 0.900 & 1.42 & 90.53 & 2.09\\
DeiT-B & 83.78 & 0.899 & 0.898 & 0.899 & 1.24 & 11.59 & 1.59\\
\addlinespace
ViT-S & 81.30 & 0.899 & 0.899 & 0.899 & 1.37 & 4.53 & 1.72\\
ViT-B & 84.61 & 0.900 & 0.900 & 0.899 & 1.19 & 4.64 & 1.54\\
\bottomrule
\end{tabular}
}
\end{subtable}
\vspace{2em}
\begin{subtable}{0.55\textwidth}
\caption{Performance on ImageNet-V2}\label{tab:INv2}

\centering
\scalebox{0.75}{
\begin{tabular}{@{}lccccccc@{}}
\toprule
  \multirow[c]{2}{*}{Model} & \multirow[c]{2}{*}{Accuracy}  &   \multicolumn{3}{c}{Coverage}    &  \multicolumn{3}{c}{Inefficiency}  \\
& & THR & APS & RAPS & THR & APS & RAPS  \\
\hline
\addlinespace
ResNet-50 & 63.15 & 0.809 & 0.863 & 0.766 & 2.46 & 15.80 & 2.25 \\
ResNet-152 & 66.95 & 0.815 & 0.861 & 0.782 & 2.08 & 11.60 & 2.01 \\
\addlinespace
DeiT-S & 70.70 & 0.811 & 0.901 & 0.840 & 1.61 & 133.04 & 2.64\\
DeiT-B & 73.28 & 0.813 & 0.877 & 0.838 & 1.344 & 18.15 & 1.89\\
\addlinespace
ViT-S & 70.32 & 0.807 & 0.872 & 0.831 & 1.48 & 8.17 & 2.09 \\
ViT-B & 74.00 & 0.805 & 0.877 & 0.843 & 1.22 & 9.10 & 1.89\\
\bottomrule
\end{tabular}
}
\end{subtable}
\begin{subtable}{0.5\textwidth}
\caption{Performance on ImageNet-R}\label{tab:INv2}
\centering
\scalebox{0.75}{
\begin{tabular}{@{}lccccccc@{}}
\toprule
  \multirow[c]{2}{*}{Model} & \multirow[c]{2}{*}{Accuracy}  &   \multicolumn{3}{c}{Coverage}    &  \multicolumn{3}{c}{Inefficiency}  \\
& & THR & APS & RAPS & THR & APS & RAPS  \\
\hline
\addlinespace
ResNet-50 & 36.16 & 0.504 & 0.710 & 0.489 & 3.37 & 26.55 & 3.49\\
ResNet-152 & 41.33 & 0.534 & 0.723 & 0.528 & 2.74 & 22.72 & 3.14 \\
\addlinespace
DeiT-S & 45.96 & 0.508 & 0.851 & 0.588 & 1.39 & 73.53 & 5.25\\
DeiT-B & 53.44 & 0.554 & 0.809 & 0.626 & 1.01 & 43.12 & 3.66\\
\addlinespace
ViT-S & 46.05 & 0.516 & 0.777 & 0.592 & 1.32 & 26.30 & 3.59\\
ViT-B & 56.84 & 0.581 & 0.836 & 0.679 & 0.95 & 27.47 & 3.20\\
\bottomrule
\end{tabular}
}
\end{subtable}
\vspace{2em}
\begin{subtable}{0.55\textwidth}
\caption{Performance on ImageNet-A}\label{tab:INv2}
\centering
\scalebox{0.75}{
\begin{tabular}{@{}lccccccc@{}}
\toprule
  \multirow[c]{2}{*}{Model} & \multirow[c]{2}{*}{Accuracy}  &   \multicolumn{3}{c}{Coverage}    &  \multicolumn{3}{c}{Inefficiency}  \\
& & THR & APS & RAPS & THR & APS & RAPS  \\
\hline
\addlinespace
ResNet-50 & 0.03 & 0.029 & 0.203 & 0.020 & 3.08 & 16.13 & 3.08\\
ResNet-152 & 5.95 & 0.176 & 0.402 & 0.165 & 3.08 & 15.86 & 3.05\\
\addlinespace
DeiT-S & 25.95 & 0.396 & 0.828 & 0.519 & 2.25 & 69.38 & 5.19 \\
DeiT-B & 38.80 & 0.469 & 0.745 & 0.591 & 1.41 & 30.66 & 3.27 \\
\addlinespace
ViT-S & 26.75 & 0.375 & 0.673 & 0.478 & 1.75 & 13.95 & 3.29 \\
ViT-B & 43.05 & 0.486 & 0.803 & 0.636 & 1.20 & 16.39 & 3.13 \\

\bottomrule
\end{tabular}
}
\end{subtable}
\begin{subtable}{0.5\textwidth}
\caption{Average performance on ImageNet-C -- contrast}\label{tab:INv2}
\centering
\scalebox{0.75}{
\begin{tabular}{@{}lccccccc@{}}
\toprule
  \multirow[c]{2}{*}{Model} & \multirow[c]{2}{*}{Accuracy}  &   \multicolumn{3}{c}{Coverage}    &  \multicolumn{3}{c}{Inefficiency}  \\
& & THR & APS & RAPS & THR & APS & RAPS  \\
\hline
\addlinespace
ResNet-50 & 35.68 & 0.512 & 0.878 & 0.528 & 2.66 & 177.81 & 4.89\\
ResNet-152 & 39.39 & 0.528 & 0.844 & 0.541 & 2.27 & 137.79 & 4.27\\
\addlinespace
DeiT-S & 69.10 & 0.795 & 0.927 & 0.842 & 1.61 & 143.79 & 3.35 \\
DeiT-B & 72.87 & 0.765 & 0.984 & 0.892 & 1.13 & 216.36 & 4.37 \\
\addlinespace
ViT-S & 55.23 & 0.626 & 0.899 & 0.719 & 1.25 & 90.40 & 3.79 \\
ViT-B & 65.09 & 0.682 & 0.913 & 0.794 & 1.04 & 82.53 & 3.20 \\
\bottomrule
\end{tabular}
}
\end{subtable}
\vspace{2em}
\begin{subtable}{0.55\textwidth}
\caption{Average performance on ImageNet-C -- brightness}\label{tab:INv2}
\centering
\scalebox{0.75}{
\begin{tabular}{@{}lccccccc@{}}
\toprule
  \multirow[c]{2}{*}{Model} & \multirow[c]{2}{*}{Accuracy}  &   \multicolumn{3}{c}{Coverage}    &  \multicolumn{3}{c}{Inefficiency}  \\
& & THR & APS & RAPS & THR & APS & RAPS  \\
\hline
\addlinespace
ResNet-50 & 64.99 & 0.827 & 0.902 & 0.797 & 2.52 & 28.38 & 2.53\\
ResNet-152 & 68.90 & 0.842 & 0.904 & 0.817 & 2.13 & 21.39 & 2.23\\
\addlinespace
DeiT-S & 76.16 & 0.860 & 0.917 & 0.883 & 1.53 & 131.98 & 2.58 \\
DeiT-B & 78.97 & 0.862 & 0.915 & 0.888 & 1.30 & 22.78 & 1.96 \\
\addlinespace
ViT-S & 75.20 & 0.869 & 0.895 & 0.876 & 1.44 & 8.34 & 2.01 \\
ViT-B & 79.27 & 0.856 & 0.899 & 0.878 & 1.22 & 8.26 & 1.78 \\
\bottomrule
\end{tabular}
}
\end{subtable}
\begin{subtable}{0.5\textwidth}
\caption{Average performance on ImageNet-C -- Gaussian noise}\label{tab:INv2}
\centering
\scalebox{0.75}{
\begin{tabular}{@{}lccccccc@{}}
\toprule
  \multirow[c]{2}{*}{Model} & \multirow[c]{2}{*}{Accuracy}  &   \multicolumn{3}{c}{Coverage}    &  \multicolumn{3}{c}{Inefficiency}  \\
& & THR & APS & RAPS & THR & APS & RAPS  \\
\hline
\addlinespace
ResNet-50 & 32.94 & 0.485 & 0.815 & 0.487 & 2.94 & 114.90 & 4.40\\
ResNet-152 & 42.10 & 0.573 & 0.839 & 0.576 & 2.55 & 85.40 & 3.64\\
\addlinespace
DeiT-S & 61.45 & 0.727 & 0.940 & 0.801 & 1.67 & 236.23 & 3.96 \\
DeiT-B & 68.77 & 0.756 & 0.945 & 0.837 & 1.28 & 84.12 & 3.03 \\
\addlinespace
ViT-S & 58.31 & 0.677 & 0.887 & 0.748 & 1.43 & 40.84 & 3.09 \\
ViT-B & 67.40 & 0.728 & 0.912 & 0.817 & 1.16 & 34.06 & 2.65 \\
\bottomrule
\end{tabular}
}
\end{subtable}
\begin{subtable}{0.55\textwidth}
\caption{Average performance on ImageNet-C -- motion blur}\label{tab:INv2}
\centering
\scalebox{0.75}{
\begin{tabular}{@{}lccccccc@{}}
\toprule
  \multirow[c]{2}{*}{Model} & \multirow[c]{2}{*}{Accuracy}  &   \multicolumn{3}{c}{Coverage}    &  \multicolumn{3}{c}{Inefficiency}  \\
& & THR & APS & RAPS & THR & APS & RAPS  \\
\hline
\addlinespace
ResNet-50 & 36.19 & 0.528 & 0.864 & 0.543 & 3.10 & 117.77 & 4.54\\
ResNet-152 & 45.20 & 0.609 & 0.873 & 0.624 & 2.51 & 80.87 & 3.68\\
\addlinespace
DeiT-S & 55.30 & 0.664 & 0.933 & 0.746 & 1.75 & 253.38 & 4.25 \\
DeiT-B & 63.01 & 0.706 & 0.897 & 0.777 & 1.34 & 43.27 & 2.73 \\
\addlinespace
ViT-S & 59.23 & 0.687 & 0.884 & 0.756 & 1.46 & 30.51 & 3.00 \\
ViT-B & 66.56 & 0.718 & 0.894 & 0.801  & 1.16 & 27.10 & 2.56 \\
\bottomrule
\end{tabular}
}
\end{subtable}

\label{tab:IN-OOD}
\end{table*}


Table \ref{tab:in-c} further highlights the brittleness of conformal prediction methods. Here, we can see that even minor corruption levels frequently lead to a violation of the target coverage. This is especially noticeable in the combination of smaller networks such as ResNet-50 and the THR method, where the smallest corruption level leads to coverage violations across all corruption types. We can also see that some corruption types lead to a greater degradation than others: motion blur tends to perform worse on average and brightness the best. In spite of the frequent degradation, the combination of DeiT-B / ViT-B and the RAPS algorithm performs consistently better across many settings, maintaining coverage levels only a few percent below the target up to corruption level 3 on most datasets.

\begin{table*}[h]
    \caption{Coverage on four different corruption types from the ImageNet-C dataset. The coverage target of 0.90 is frequently violated even with minor levels of corruption.}
\centering
\scalebox{0.65}{
    \begin{tabular}{@{}lcccccc|ccccc|ccccc|ccccc@{}}
    \toprule
      \multirow[c]{2}{*}{Model} & \multirow[c]{2}{*}{CP Method} & \multicolumn{5}{c}{Contrast}  &   \multicolumn{5}{c}{Brightness}    &  \multicolumn{5}{c}{Gaussian Blur}  &  \multicolumn{5}{c}{Motion Blur}  \\
    &  & 1 & 2 & 3 &4 & 5 & 1 & 2 & 3 &4 & 5& 1 & 2 & 3 &4 & 5& 1 & 2 & 3 &4 & 5  \\
    \hline
    \addlinespace
     \multirow[c]{2}{*}{ResNet-50} & THR & 0.808 & 0.753 & 0.627 & 0.300 & 0.073 & 0.871 & 0.860 &  0.840 & 0.809 & 0.756 & 0.779 & 0.687 & 0.524 & 0.314 & 0.115 & 0.809 & 0.714 & 0.533 &0.339 & 0.245\\
     & RAPS & 0.780 & 0.734 & 0.632 & 0.360 & 0.137 & 0.832 & 0.822 & 0.807 &0.783 & 0.741 & 0.845 & 0.797 & 0.692 & 0.538 & 0.326& 0.786 & 0.711 & 0.553 & 0.378 & 0.288  \\
    \addlinespace
    \addlinespace
     \multirow[c]{2}{*}{ResNet-152} & THR & 0.823 & 0.773 & 0.657 & 0.320 & 0.066 & 0.880 & 0.871 & 0.854 & 0.826 & 0.780 & 0.808 & 0.747 & 0.627 & 0.444 & 0.213 & 0.852 & 0.771 & 0.638 & 0.460 & 0.340  \\
     & RAPS & 0.800 & 0.760 & 0.661 & 0.368 & 0.117 & 0.847 & 0.839 & 0.827 & 0.804 & 0.768 & 0.862 & 0.829 & 0.767 &0.642 & 0.432 & 0.882 & 0.763 & 0.652 &0.500 & 0.390  \\
    \addlinespace
    \addlinespace
     \multirow[c]{2}{*}{DeiT-S} & THR & 0.867 & 0.859 & 0.840 & 0.778 & 0.633 & 0.881 & 0.876 & 0.867 & 0.851 & 0.825 & 0.853 & 0.829 & 0.772 &0.673 & 0.497 & 0.852 & 0.802 & 0.698 &0.540 & 0.429  \\
     & RAPS & 0.887 & 0.880 & 0.870 &0.835 & 0.735 & 0.893 & 0.891 & 0.887 & 0.879 & 0.865 & 0.885 & 0.872 & 0.838 & 0.777 & 0.648 & 0.882 & 0.853 & 0.783 &0.656 & 0.559  \\
    \addlinespace
    \addlinespace
     \multirow[c]{2}{*}{DeiT-B} & THR & 0.853 & 0.845 & 0.825 & 0.750 & 0.551 & 0.880 & 0.878 & 0.868 & 0.854 & 0.834 & 0.857 & 0.836 & 0.793 & 0.722 & 0.578 & 0.851 & 0.814 & 0.739 &0.612 & 0.513  \\
     & RAPS & 0.930 & 0.927 & 0.919 & 0.887 & 0.797 & 0.894 & 0.893 & 0.892 & 0.887 & 0.874 & 0.887 & 0.886 & 0.869 & 0.828 & 0.738 & 0.873 & 0.851 & 0.801 &0.709 & 0.630  \\
    \addlinespace
    \addlinespace
     \multirow[c]{2}{*}{ViT-S} & THR & 0.844 & 0.820 & 0.761 &0.542 & 0.163 & 0.877 & 0.871 & 0.859 & 0.838 & 0.805 & 0.842 & 0.807 & 0.737 & 0.609 & 0.381 & 0.845 & 0.803 & 0.716 & 0.582 & 0.488  \\
     & RAPS & 0.869 & 0.856 & 0.828 &0.697 & 0.349 & 0.886 & 0.882 & 0.874 & 0.861 & 0.841 & 0.867 & 0.849 & 0.802 & 0.714 & 0.535& 0.868 & 0.843 & 0.785 & 0.688 & 0.610  \\
    \addlinespace
    \addlinespace
     \multirow[c]{2}{*}{ViT-B} & THR & 0.853 & 0.838 & 0.801 & 0.637 & 0.280 & 0.877 & 0.872 & 0.863 & 0.846 & 
 0.822 & 0.848 & 0.821 & 0.771 & 0.683 & 0.523 & 0.852 & 0.819 & 0.747 & 0.632 & 0.542  \\
     & RAPS & 0.880 & 0.874 & 0.862 & 0.804 & 0.551 & 0.888 & 0.886 & 0.881 & 0.874 & 0.862 & 0.885 & 0.878 & 0.856 & 0.807 & 0.696 & 0.881 & 0.864 & 0.824 & 0.752 & 0.684  \\

    \bottomrule
    \end{tabular}   
}
    \label{tab:in-c}
\end{table*}

\section{Ineffectiveness of class-balanced CP on PlantNet-300k}

As demonstrated in Table \ref{tab:pn300k}, performing conformal prediction on long-tailed data leads to large violations in class-conditional coverage. Class-conditional coverage can be approached through \textbf{class balanced} conformal prediction, which aims to ensure that the specified error rates are guaranteed for \textit{every} class. \citet{sadinle_least_2019} propose a method that calibrates a threshold for each class, then including classes in the confidence set based on their class-specific thresholds:
\begin{equation}
    C(x;\tau) = \left\{k \in [K] : s(x,k) <\tau^{(k)} \right\}.
\end{equation}
This method can be used in conjunction with the other post-hoc CP methods described in Appendix \ref{confpred_apdx}.

We investigated class-conditional conformal prediction on PlantNet-300k, and summarize the results in Table \ref{tab:pn300k-cb}. It results in better macro-coverage and fewer coverage violations than the regular conformal prediction, yet it still leads to class-conditional coverage violations. This is partly because CP coverage holds in expectation across an infinite test set. Where ample data per class is available, like on ImageNet, this can be simulated by repeated random data splits. PlantNet-300k has fixed calibration / test sets and as noted, some classes have very little representation. Further, coverage follows a Beta distribution with $\alpha$ and $\beta$ terms reliant on the validation set size \cite{vovk2012conditional}, thus a smaller calibration set leads to greater variance in coverage across the (infinite) test set. Thus, when class-balanced conformal prediction is performed on PlantNet-300k, both \textit{the calibration and test sets for each class are very small} due to the long-tailed label distribution. This leads to a high class-conditional variance in coverage and thus does not resolve the coverage violations. \par
Although this is a challenging setting, it is nonetheless reflective of possible scenarios that can be encountered in the real world. One may imagine many data-constrained environments such as medicine where gathering a large number of examples for rare (yet still important) classes is a challenging feat. If conformal prediction is to be deployed in these settings, this is a hurdle that must be addressed.  

\begin{table*}[t]

\caption{Class-balanced conformal prediction results on PlantNet-300k. Although this leads to better macro-coverage results than regular (marginal) CP, class-conditional coverage is still violated due to the high coverage variance associated with the small, and fixed, per-class calibration / test set sizes}
\centering
\scalebox{0.82}{
    \begin{tabular}{@{}lcccccc@{}}
    \toprule
      Model & CP Method  & Coverage & Macro Coverage & Inefficiency & Macro Inefficiency & \# (\%)~Violated Classes \\
    \hline
    \addlinespace
     \multirow[c]{3}{*}{ResNet-152} & THR & 0.920 & 0.676 & 12.66 & 29.10 & 460~(43\%)\\
     & APS & 0.998 & 0.983 & 643.50 & 737.12 & 24~(2\%) \\
     & RAPS &  0.978 & 0.802 & 25.10 & 27.93 & 268~(25\%) \\
    \addlinespace
     \multirow[c]{3}{*}{DeiT-B} & THR & 0.901 & 0.747 & 13.56 & 22.36 & 372~(34\%) \\
     & APS & 0.998 & 0.979 & 689.90 & 682.48 & 27~(2\%) \\
     & RAPS  & 0.968 & 0.836 &16.00 & 22.47 & 259~(24) \\
    \addlinespace
     \multirow[c]{3}{*}{ViT-B} & THR& 0.916 & 0.659 & 10.60 & 43.25 & 480~(44\%) \\
     & APS &  0.991 & 0.951 & 331.80 & 656.73 & 76~(7\%) \\
     & RAPS & 0.966 & 0.951 & 28.75 & 30.67 & 360~(33\%)\\
     \bottomrule
    \end{tabular}   
}
\label{tab:pn300k-cb}
\end{table*}

\section{iNaturalist Results}

The iNaturalist-2018 \cite{inaturalist18} and iNaturalist-2019 \cite{inaturalist19} datasets both feature long-tailed training sets and class-balanced test sets. They are comprised of 8,142 and 1,010 classes, respectively. Here, 50\% of the test set is used to calibrate the conformal threshold, and the remainder is used to predict confidence sets. Unlike the PlantNet-300k dataset, the conformal calibration process is conducted on a class-balanced dataset. We can see in Table \ref{tab:inat} that this results in a considerably lower percentage of classes with violated coverage. 

\begin{table*}[t]

\caption{Conformal prediction on iNat-2018 (\subref{tab:inat2018}) and iNat-2019 (\subref{tab:inat2019}). Although the conformal thresholds are calibrated on a class-balanced dataset, there is frequent violation of class-conditional coverage. }

\begin{subtable}{\textwidth}
\caption{Conformal prediction results on iNat-2018. }
\centering
\scalebox{0.85}{
    \begin{tabular}{@{}lccccc@{}}
    \toprule
      Model & CP Method & Accuracy & Coverage & Inefficiency & Num. (\%)~Violated Classes \\
    \hline
    \addlinespace
     \multirow[c]{3}{*}{ResNet-152} & THR &\multirow[c]{2}{*}{50.31} & 0.901 & 16.39 & 1,074~(13\%) \\
     & RAPS & & 0.905 & 16.58 & 1,078~(13\%) \\
    \addlinespace
     \multirow[c]{3}{*}{DeiT-B} & THR &\multirow[c]{2}{*}{74.66} & 0.905 & 2.82 & 1,105~(14\%) \\
     & RAPS & & 0.907 & 3.47 & 1,093~(13\%) \\
    \addlinespace
     \multirow[c]{3}{*}{ViT-B} & THR &\multirow[c]{2}{*}{61.61} & 0.899 & 7.19 & 1,119~(14\%) \\
     & RAPS & & 0.899 & 9.15 & 1,125~(14\%) \\
     \bottomrule
    \end{tabular}   
}
\label{tab:inat2018}
\end{subtable}

\hspace{1mm}
\newline
\begin{subtable}{\textwidth}
\caption{Conformal prediction results on iNat-2019.}

\centering
\scalebox{0.85}{
    \begin{tabular}{@{}lccccc@{}}
    \toprule
      Model & CP Method & Accuracy & Coverage & Inefficiency & Num. Violated Classes \\
    \hline
    \addlinespace
     \multirow[c]{3}{*}{ResNet-152} & THR &\multirow[c]{2}{*}{62.97} & 0.900 & 4.01 & 113~(11\%) \\
     & RAPS & & 0.900 & 5.31 & 147~(15\%) \\
    \addlinespace
     \multirow[c]{3}{*}{DeiT-B} & THR &\multirow[c]{2}{*}{78.42} & 0.899 & 1.58 & 140~(14\%) \\
     & RAPS & & 0.900 & 2.03 & 141~(14\%) \\
    \addlinespace
     \multirow[c]{3}{*}{ViT-B} & THR &\multirow[c]{2}{*}{75.71} & 0.900 & 2.00 & 146~(14\%) \\
     & RAPS & & 0.904 & 2.29 & 146~(14\%) \\
     \bottomrule
    \end{tabular}   
}
\label{tab:inat2019}

\end{subtable}
\label{tab:inat}
\end{table*}

\section{The Relationship between Accuracy and Coverage}\label{apndx:acc_vs_cov}
\begin{figure*}
  \begin{subfigure}{\linewidth}
  \includegraphics[width=.5\linewidth]{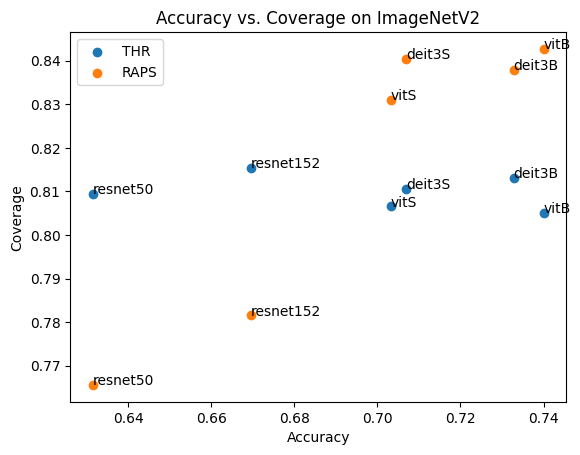}
  \includegraphics[width=.5\linewidth]{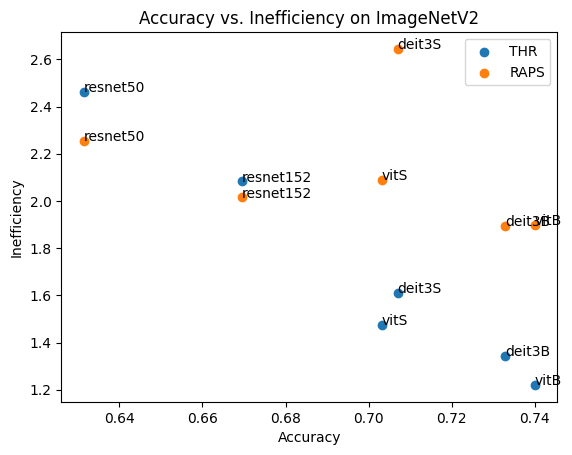}
  \caption{Coverage (left) and Inefficiency (right) on ImageNetV2}
  \end{subfigure}\par\medskip
  \begin{subfigure}{\linewidth}
  \includegraphics[width=.5\linewidth]{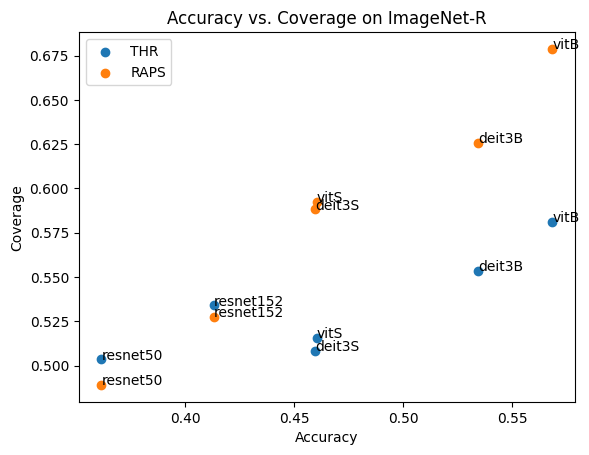}
  \includegraphics[width=.5\linewidth]{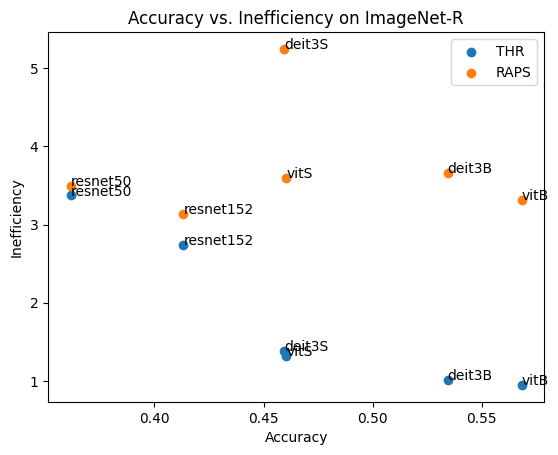}
  \caption{Coverage (left) and Inefficiency (right) on ImageNet-R}
  \end{subfigure}\par\medskip
  \begin{subfigure}{\linewidth}
  \includegraphics[width=.5\linewidth]{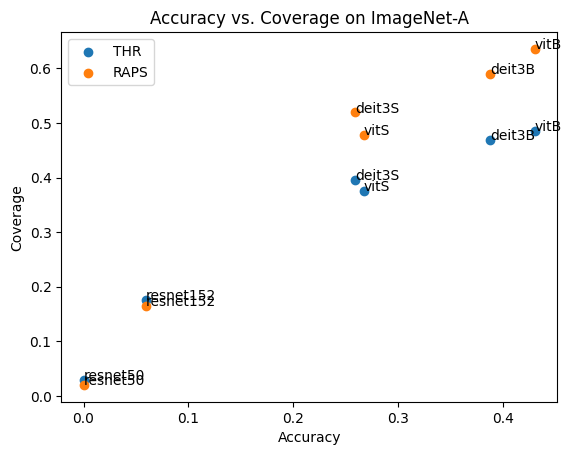}
  \includegraphics[width=.5\linewidth]{images/inef_inv2.png}
  \caption{Coverage (left) and Inefficiency (right) on ImageNet-A}
  \end{subfigure}
  \caption{An increase in coverage and inefficiency performance is generally followed by an increase in the accuracy of the underlying model. The target coverage is 0.90 across all datasets. }
  \label{fig:acc_vc_cov}
\end{figure*}

Figure \ref{fig:acc_vc_cov} plots the relation between coverage / inefficiency performance and the accuracy of the underlying model, on on the different distribution-shifted datasets. We can observe that coverage generally increases along with accuracy. Inefficiency also improves, albeit the THR method seems to have larger inefficiency improvements. \par

\begin{figure*}
  \begin{subfigure}{\linewidth}
  \includegraphics[width=.5\linewidth]{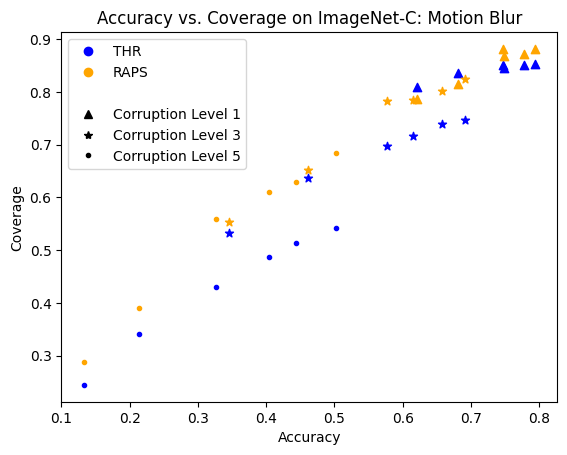}
  \includegraphics[width=.5\linewidth]{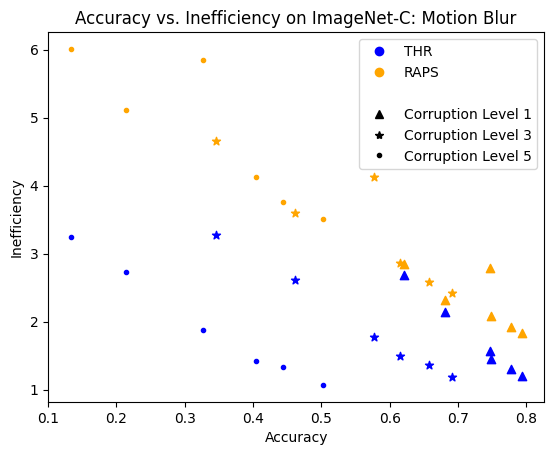}
  \caption{Coverage (left) and Inefficiency (right) on ImageNet-C --- motion blur}
  \end{subfigure}\par\medskip
  \begin{subfigure}{\linewidth}
  \includegraphics[width=.5\linewidth]{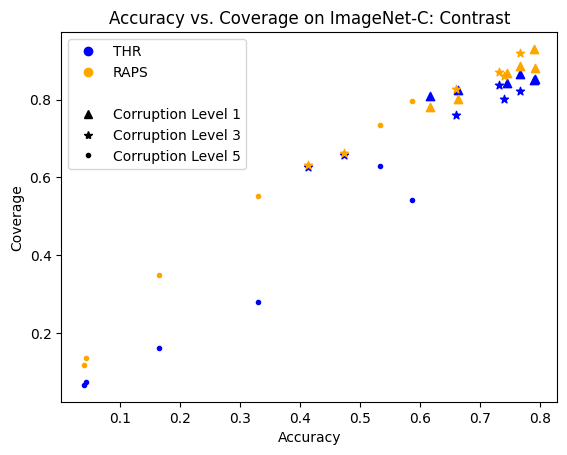}
  \includegraphics[width=.5\linewidth]{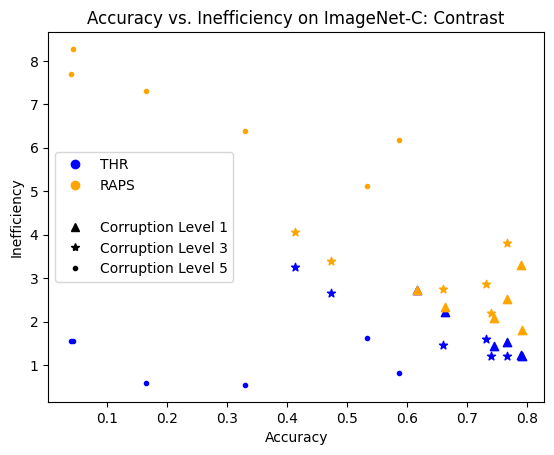}
  \caption{Coverage (left) and Inefficiency (right) on ImageNet-C --- contrast}
  \end{subfigure}\par\medskip
  \begin{subfigure}{\linewidth}
  \includegraphics[width=.5\linewidth]{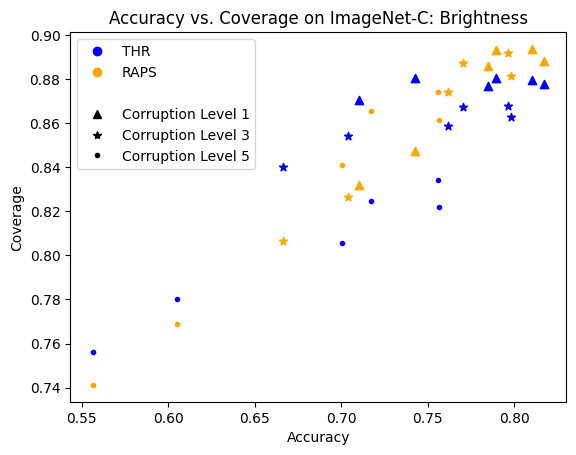}
  \includegraphics[width=.5\linewidth]{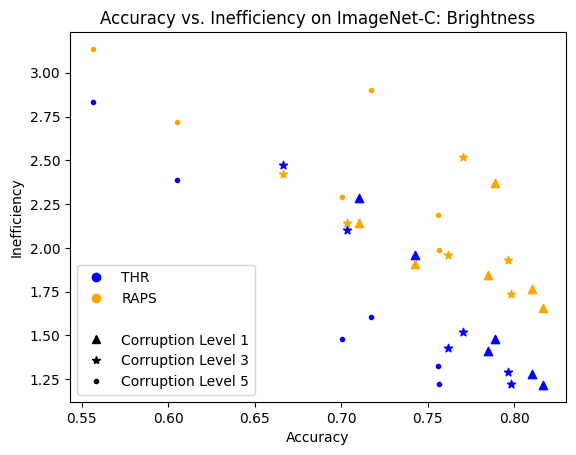}
  \caption{Coverage (left) and Inefficiency (right) on ImageNet-C --- brightness}
  \end{subfigure}
  \caption{Coverage / inefficiency vs.~accuracy for various levels of corruption. The target coverage is 0.90.}
  \label{fig:acc_cov_inC}
\end{figure*}

Similarly, Figure \ref{fig:acc_cov_inC} plots the coverage / inefficiency relation with accuracy for various corruption levels and types. There is a marked improvement in coverage when the underlying model is more accurate, which seems especially pronounced for greater levels of corruption. \par


Interestingly, the relation between the accuracy of the underlying neural network and coverage / inefficiency appears to vary with the CP method used. For example, RAPS generally demonstrates a near linear increase in coverage with increased accuracy, however inefficiency gains seem to diminish. Conversely, the inefficiency of the THR algorithm consistently improves with accuracy, and coverage gains are less pronounced. 

\section{Results on ImageNet-W}\label{apndx:INw}

Recent work \cite{li_2023_whac_a_mole} has found a reliance on translucent watermarks as a shortcut in current vision models, and the addition of a watermark on the ImageNet validation set leads to large decreases in performance. We investigate this dataset, called ImageNet-W, in the conformal prediction setting and similarly find a general decrease in coverage across most models and methods. As seen in Table \ref{tab:IN-W}, the APS method combined with vision transformers is able to maintain coverage on this dataset, at the expense of larger set sizes. This reemphasizes both the brittleness of some conformal prediction methods; a simple watermark is sufficient in violating coverage guarantees, as well as the potential for improvement using better deep learning models and different CP methods.

\begin{table}[h]
    \caption{Coverage and inefficiency when calibrating on ImageNet and evaluating on ImageNet-W \cite{li_2023_whac_a_mole}. The target coverage is 0.90. }
\centering
\begin{tabular}{@{}lccccccc@{}}
\toprule
  \multirow[c]{2}{*}{Model} & \multirow[c]{2}{*}{Accuracy}  &   \multicolumn{3}{c}{Coverage}    &  \multicolumn{3}{c}{Inefficiency}  \\
    & & THR & APS & RAPS & THR & APS & RAPS  \\
    \hline
    \addlinespace
    ResNet-50 & 48.66 & 0.657 & 0.734 & 0.621 & 3.07 & 12.77 & 2.65 \\
    ResNet-152 & 48.54 & 0.637 & 0.741 & 0.617 & 2.83 & 12.21 & 2.74 \\
    \addlinespace
    DeiT-S & 75.41 & 0.851 & 0.904 & 0.872 & 1.52 & 111.13 & 2.53 \\
    DeiT-B & 78.39 & 0.855 & 0.909 & 0.880 & 1.28 & 22.36 & 1.91 \\
    \addlinespace
    ViT-S & 73.86 & 0.842 & 0.909 & 0.873 & 1.47 & 9.20 & 2.21 \\
    ViT-B & 79.30 & 0.855 & 0.912 & 0.888 & 1.22 & 9.01 & 1.87\\
    \bottomrule
    \end{tabular} 
    
    \label{tab:IN-W}
\end{table}


\end{document}